\documentclass[letterpaper, 10 pt, conference]{ieeeconf}  

\IEEEoverridecommandlockouts                              

\usepackage[detect-none,binary-units=true]{siunitx}
\DeclareSIUnit{\nothing}{\relax}
\usepackage{graphicx}
\usepackage{footnote} 
\usepackage{hyperref}
\usepackage{booktabs}
\usepackage{multirow}

\hypersetup{%
    pdfborder = {0 0 0}
}
\usepackage{color}
\usepackage{tabularray}
\usepackage{array}
\usepackage{pifont}
\newcommand{\cmark}{\ding{51}}%
\newcommand{\xmark}{\ding{55}}%

\usepackage{xfrac}

\DeclareSIUnit\op{Op}

\definecolor{somegray}{rgb}{0.5, 0.5, 0.5}
\newcommand{\darkgrayed}[1]{\textcolor{somegray}{#1}}
\makeatletter
\newcommand*\titleheader[1]{\gdef\@titleheader{#1}}
\AtBeginDocument{%
  \let\st@red@title\@title
  \def\@title{%
    \vskip-3.0em
    \bgroup\normalfont\large\centering\@titleheader\par\egroup
    \vskip0.0em\st@red@title}
}

\makeatother

\titleheader{\darkgrayed{This paper has been accepted for publication in the IEEE IROS 2023 conference\\\copyright 2023 IEEE.}}

\title{\LARGE \bf
Sim-to-Real Vision-depth Fusion CNNs for Robust Pose Estimation Aboard Autonomous Nano-quadcopters}

\author{Luca Crupi$^{1}$, Elia Cereda$^{1}$, Alessandro Giusti$^{1}$, and Daniele Palossi$^{12}$
\thanks{This work was partially supported by the Swiss National Science Foundation (SNSF) through the NCCR Robotics.}
\thanks{$^{1}$L. Crupi, E. Cereda, A. Giusti, and D. Palossi are with the Dalle Molle Institute for Artificial Intelligence (IDSIA), USI and SUPSI, Lugano, 6962, Switzerland
        {\tt\small name.surname@idsia.ch}}%
\thanks{$^{2}$D. Palossi is also with the Integrated Systems Laboratory (IIS), ETH Z\"urich, Z\"urich, 8092, Switzerland
        {\tt\small dpalossi@iis.ee.ethz.ch}}%
}

\begin{document}

\maketitle
\thispagestyle{empty}
\pagestyle{empty}

\begin{abstract}
Nano-quadcopters are versatile platforms attracting the interest of both academia and industry.
Their tiny form factor, i.e., $\sim$\SI{10}{\centi\meter} diameter, makes them particularly useful in narrow scenarios and harmless in human proximity.
However, these advantages come at the price of ultra-constrained onboard computational and sensorial resources for autonomous operations.
This work addresses the task of estimating human pose aboard nano-drones by fusing depth and images in a novel CNN exclusively trained in simulation yet capable of robust predictions in the real world.
We extend a commercial off-the-shelf (COTS) Crazyflie nano-drone --- equipped with a 320$\times$240 px camera and an ultra-low-power System-on-Chip --- with a novel multi-zone (8$\times$8) depth sensor.
We design and compare different deep-learning models that fuse depth and image inputs.
Our models are trained exclusively on simulated data for both inputs, and transfer well to the real world: field testing shows an improvement of 58\% and 51\% of our depth+camera system w.r.t. a camera-only State-of-the-Art baseline on the horizontal and angular mean pose errors, respectively.
Our prototype is based on COTS components, which facilitates reproducibility and adoption of this novel class of systems.
\end{abstract}

\section*{Supplementary video material}
In-field tests: \href{https://youtu.be/p4s2j0_6828}{https://youtu.be/p4s2j0\_6828}.

\section{Introduction} \label{sec:intro}

Miniaturized autonomous quadcopters (\textit{nano-drones}) are extending the application areas of aerial robotics, from exploration in narrow spaces~\cite{dumbgen2022blind} to human-robot interaction~\cite{pulp-frontnet}.  
With a diameter of $\sim$\SI{10}{\centi\meter}, nano-drones can reach inaccessible places for bigger flying robots and safely operate near humans.
Additionally, nano-drone hardware is relatively cheap compared to bigger and more powerful multi-rotors, making these platforms even more attractive.

Still, these platforms come at the price of extremely limited onboard resources, such as memories, processors, and sensors~\cite{DCOSS19}.
This is a significant drawback in comparison to standard-sized drones, with a diameter of $\sim$\SI{50}{\centi\meter} and a few \SI{}{\kilo\gram} of payload, that can cope with complex environments and sophisticated workloads~\cite{scaramuzza_alphapilot,deeppilot_2020} thanks to onboard powerful processors and rich sensors, such as LIDAR~\cite{uav-lidar} and depth cameras~\cite{uav-rgbd}.  
As an example, the GWT GAP8 System-on-Chip (SoC) that equips our nano-drone platform peaks at \SI{22}{\giga OP/\second}: three orders of magnitude less than the NVIDIA Jetson Xavier AGX flight computer, which achieves up to \SI{32}{\tera OP/\second}.

\begin{figure}[t]
\centering
\includegraphics[width=1.0\linewidth]{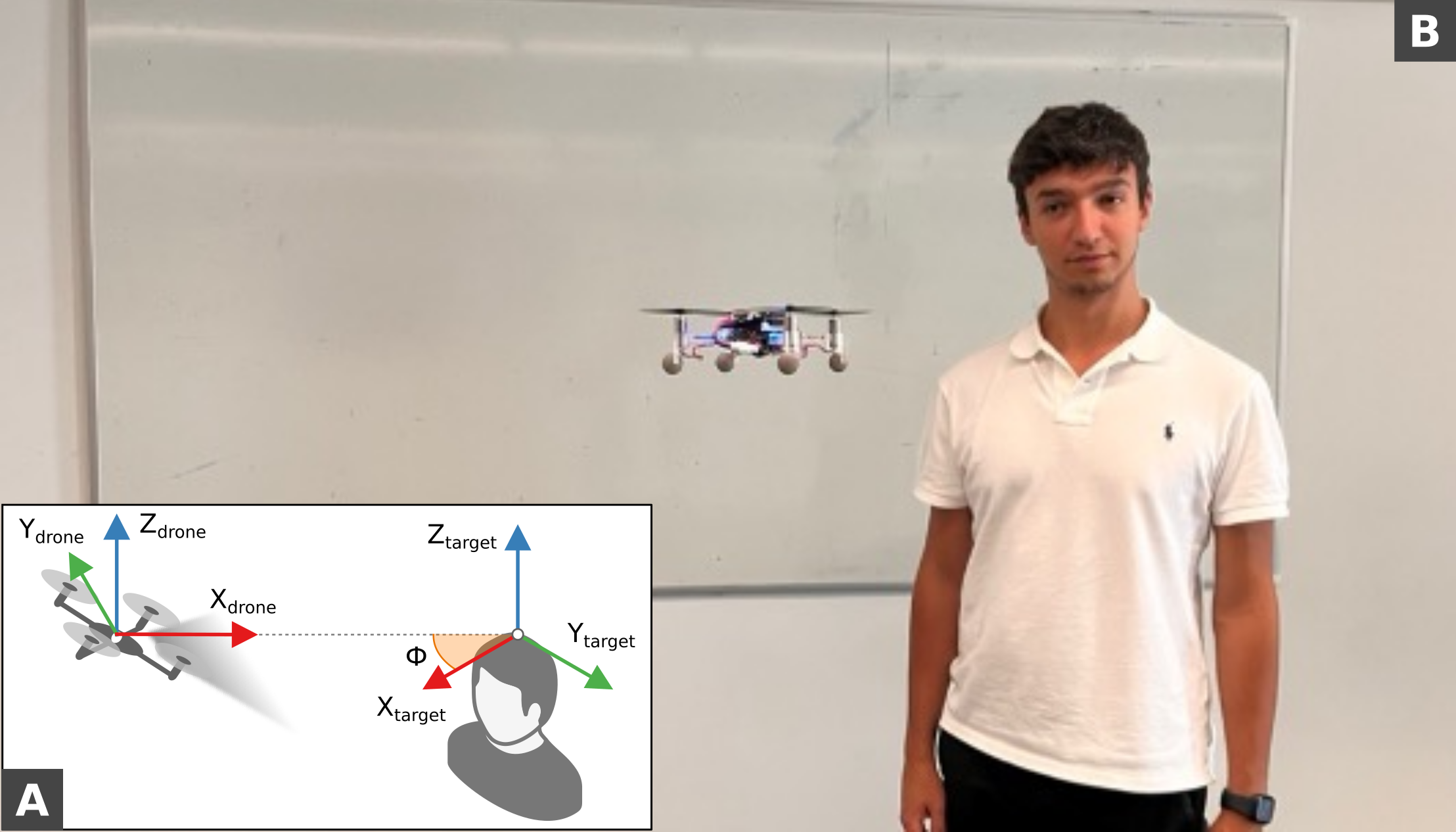}
\caption{A) Definition of our 
person pose estimation task. B) Example of person pose estimation scenario.}
\label{fig:intro}
\end{figure}

When designing nano-drone systems, simple convolutional neural network-based (CNN) approaches are an attractive solution to perception tasks since these models can be used for inference with limited resources. 
Learning-based approaches are suitable for fusing sensory streams from different onboard sensors~\cite{fusenet} since one can learn to manage noisy, low-resolution data and even capture and exploit correlations across sensors.
Different types of sensing technologies can complement each other, such as Time-of-Flight (ToF) depth sensors, which are very accurate at a low distance (a few meters), with noisy but longer-range CMOS cameras.

One challenge which arises when adopting this approach is collecting the data to train multi-sensor perception models.
Existing large image-only datasets can not be used to train multi-modal CNN since they miss the corresponding inputs from the additional sensors. 
One solution would be acquiring new data (with all sensors), which is expensive and time-consuming if not unpracticable for those cases requiring additional ad-hoc infrastructure such as an external motion capture system (mocap)~\cite{10160586}.
An option to extend existing image-only datasets with the missing sensory data can be by developing software models/pipelines to generate additional synthetic samples. 
However, this strategy is only sometimes applicable and can easily suffer from inaccuracy.
A third solution, also used in our work, relies on photorealistic simulators, which can provide abundant data from multiple sensors with little effort and time~\cite{10160586}.

\begin{table*}[t]
\centering
\caption{SoA comparison: Pose estimation task on UAVs.}
\label{tab:soa_comparison}
\begin{tblr}{
  width = \linewidth,
  colspec = {Q[60]Q[80]Q[60]Q[80]Q[120]Q[140]Q[80]Q[90]Q[120]Q[80]Q[110]},
  cells = {c},
  hline{1,7} = {-}{0.08em},
  hline{2} = {-}{0.05em},
}
\textbf{Work} & \textbf{Pose} & \textbf{Size} & \textbf{Onboard} & \textbf{Sensors} & \textbf{Resolution} & \textbf{Algorithm} & \textbf{Training} & \textbf{Device} & \textbf{Power} & \textbf{Field tested}\\
\cite{jung2018perception} & object & standard & yes & mono camera & VGA & CNN & real & Jetson~TX2 & \SI{20}{\watt} & \cmark\\
\cite{scaramuzza_alphapilot} & object & standard & yes & stereo~camera & $2\times$ 720p & CNN+geom. & real & Jetson~Xavier & \SI{20}{\watt} & \cmark\\
\cite{dronecap} & human & micro & no & mono~camera & 1080p & CNN+geom. & real & GTX~Titan~X & \SI{100}{\watt} & \xmark\\
\cite{pulp-frontnet} & human & nano & yes & mono~camera & QVGA & CNN & real & GWT~GAP8 & \SI{100}{\milli\watt} & \cmark\\
\textbf{Ours} & \textbf{human} & \textbf{nano} & \textbf{yes} & \textbf{depth+camera} & \textbf{8$\times$8 / QVGA} & \textbf{CNN} & \textbf{sim} & \textbf{GWT~GAP8} & \textbf{\SI{100}{\milli\watt}} & \textbf{\cmark}
\end{tblr}
\end{table*}

Combining \textit{i}) a commercial off-the-shelf (COTS) Bitcraze Crazyflie 2.1 nano-drone, \textit{ii}) an AI-deck expansion board, hosting a parallel ultra-low-power GWT GAP8 SoC~\cite{gap8} and a QVGA monochrome camera, and \textit{iii}) an STM 8$\times$8 multi-zone depth sensor, we introduce and demonstrate our vertically integrated system.
Our work addresses estimating the relative pose of a human w.r.t. a nano-drone, as displayed in Figure~\ref{fig:intro}, employing a CNN that fuses data from the two onboard sensors: depth sensor and monocular camera.

We contribute with \textit{i}) the design and thorough analysis of multiple CNN models fed with the two complementary sensory inputs; \textit{ii}) a detailed description of our sim-to-real pipeline, which exploits aggressive photometric augmentations and balanced label distributions; \textit{iii}) comprehensive in-field experimental results, challenging our system in real-world conditions and comparing it across various configurations, including a State-of-the-Art (SoA) baseline~\cite{pulp-frontnet}.

In-field experiments demonstrate \textit{i}) the robustness of our sim-to-real training method; \textit{ii}) the efficiency of the CNN fusing depth and images, which runs aboard the GAP8 SoC up to \SI{45}{frame/\second} within a power envelope of \SI{92}{\milli\watt}; \textit{iii}) the predictive performance of the fusion approach. 
In particular, on a never-seen-before flying arena, our system significantly outperforms a camera-only SoA baseline: the Mean Absolute Error of the estimated human position and relative orientation angle are reduced respectively by 58\% and 51\%. 
Finally, our prototype employs only ready-to-use COTS electronic components to ease the reproducibility and adoption of this novel class of systems.
\section{Related Work} \label{sec:related}

Standard/Micro-sized UAVs (50-\SI{30}{\centi\meter} diameter) heavily leverage multiple types of sensing devices to address various perception tasks.
For example, autonomous drone races, in which a precise understanding of the world is fundamental to navigate at high speeds, exploit depth information for various tasks, including estimating 3D relative poses of gates~\cite{scaramuzza_alphapilot,jung2018perception} and supporting visual-inertial odometry pipelines~\cite{jung2018perception}. 
Similarly, swarm operations of micro-sized drones can exploit depth sensors for mapping and localization tasks~\cite{zhou2022swarm}.
Typical depth sensors used in aerial robotics~\cite{onboard-adr-survey} include stereo cameras like the ZED and Intel RealSense D430 and RGB-D cameras based on ToF technology, such as the Azure Kinect.
All these accurate sensing devices come with a form factor, weight ($>$\SI{100}{\gram}), and power consumption ($>$\SI{5}{\watt}), which prevent them from being also adopted on nano-UAVs.
For this reason, aboard our nano-drone, we take advantage of the ultra-compact STM VL53LC5CX 8$\times$8 ToF-based depth sensor (sub-centimeter size and sub-gram weight) and an Himax HM01B0 QVGA camera.

Focusing on our target pose estimation task, CNNs represent a common solution to tackle this problem~\cite{scaramuzza_alphapilot,deeppilot_2020,jung2018perception,dronecap}, also on nano-drones~\cite{pulp-frontnet}.
Sophisticated multi-cameras pipelines can combine CNN-based approaches with geometric ones~\cite{scaramuzza_alphapilot}, but at the price of heavier computation.
In contrast, multi-modal CNNs directly fuse vision and depth in a unified deep-learning model~\cite{fusenet,rgbd-cnn}, ensuring the resulting models take full advantage of both modalities~\cite{input-dropout}. 
This approach is particularly relevant for our case as we can afford only an extremely low-resolution depth map aboard the nano-drone.
Therefore, in this work, we explore multi-modal CNN to maximize the information extracted from the depth map.

Despite the limited computational capabilities of microcontroller units (MCUs) aboard nano-drones, both monocular~\cite{DCOSS19,nanoflownet,nano-learning-vo,idsia-d2d,pulp-frontnet} and stereo~\cite{dewagter14-flapstereo,mcguire17-of} vision-based solutions have been proposed.
However, this last group of solutions introduces a significant computational load to accomplish relatively low-level control tasks (stabilization and obstacle avoidance).
Among these works, the PULP-Frontnet~\cite{pulp-frontnet} CNN tackles the human pose estimation aboard a Crazyflie nano-drone employing a monocular camera with real-time performance.
Starting from this SoA model, which reaches up to 48 frames per second on our platform, we present our novel multi-modal CNN, which fuses images and depth maps by running directly aboard our nano-drone. 
The constraint posed by the platform in terms of computational capabilities and by the task in terms of minimum frame rate for the in-field tracking, pose huge limits on the type of networks that can be deployed onboard.
Networks such as Resnet-18~\cite{he2015deep} and transformers~\cite{NIPS2017_3f5ee243}, with several billion multiply-accumulate operations (MACs) per frame~\cite{Liu_2022_CVPR}, are unsuitable for real-time performance on our platform.
Table~\ref{tab:soa_comparison} summarizes the aforementioned works, comparing them to our proposed approach regarding sensor modalities, adopted algorithms, and computation platforms.
\section{Background} \label{sec:background}

\begin{figure}[t]
  \includegraphics[width=\columnwidth]{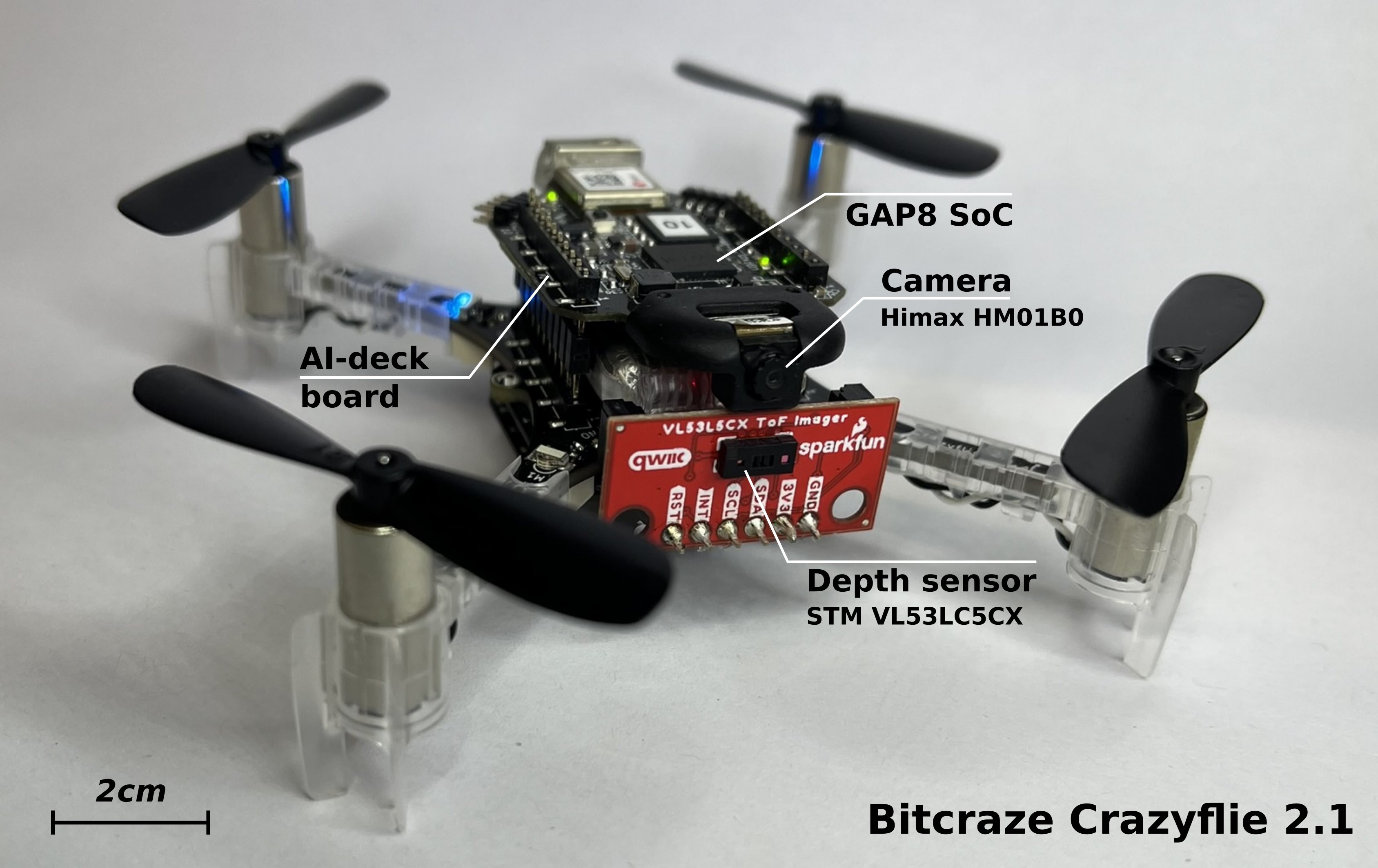}
  \caption{Our prototype is based on COTS components: the Crazyflie 2.1 nano-drone, the AI-deck board, and an STM VL53LC5CX depth sensor.}
  \label{fig:proto}
\end{figure}

\begin{figure*}[t]
  \includegraphics[width=\textwidth]{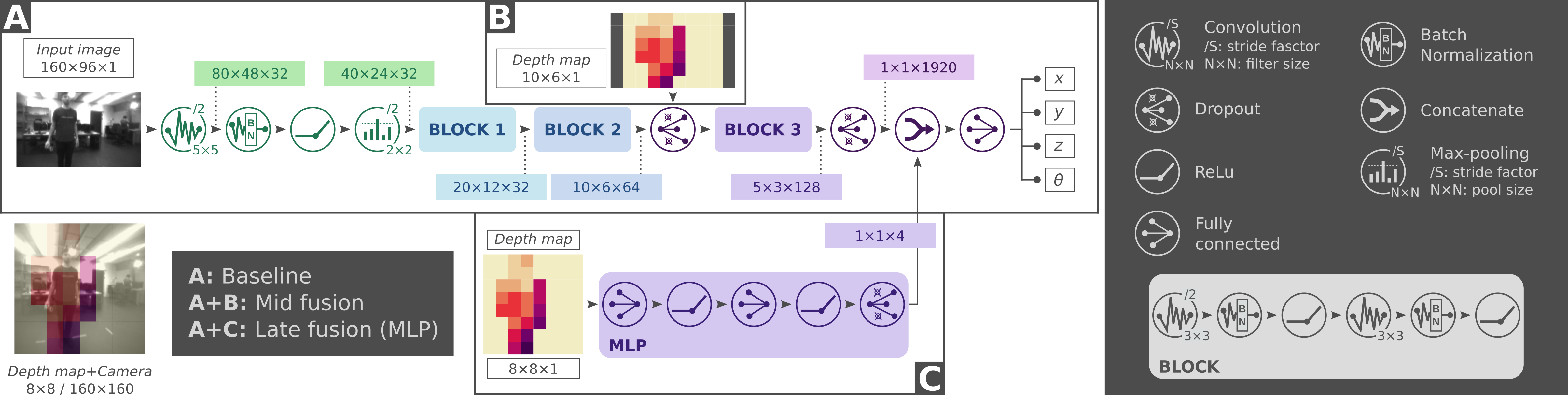}
  \caption{Our CNNs exploration. A) the SoA baseline~\cite{pulp-frontnet} used as the backbone of our Depth+Cam fusion. B) Mid-fusion model. C) Late-fusion model.}
  \label{fig:models}
\end{figure*}

\textbf{Robotic platform:} our prototype, shown in Figure~\ref{fig:proto}, is based on a Crazyflie 2.1 nano-quadrotor, an open-source \SI{27}{\gram} palm-sized COTS nano-drone.
An STM32 MCU performs low-level flight control tasks, i.e., state estimation and proportional–integral–derivative (PID) cascade controller. 
The nano-drone is extended with an AI-deck expansion board tasked with onboard high-level intelligence.
The AI-deck features a GWT GAP8 multi-core RISC\_V-based SoC, two-level off-chip memories, i.e., \SI{8}{\mega\byte} DRAM and \SI{64}{\mega\byte} Flash, and a monocular QVGA Himax HM01B0 @ \SI{60}{\hertz}.

The GAP8 SoC is divided into two power domains: a single-core \textit{fabric controller} which orchestrates accesses to external memories/sensors, and an eight-core \textit{cluster} domain optimized for parallel computation of compute-intense workloads, e.g., CNNs. 
The on-chip memory hierarchy is composed of a \SI{64}{\kilo\byte} low-latency L1 memory shared among the cluster cores and a \SI{512}{\kilo\byte} L2 memory.
The GAP8 also features two DMA engines that efficiently automate data transfers between memory levels and external peripherals, such as the UART interface, which connects the GAP8 to the STM32. 
The lack of data caches and floating-point units requires, respectively, explicit data management in software and the use of integer-quantized arithmetic.

We complement our platform with the multi-zone ranging sensor VL53LC5CX\footnote{\href{https://www.st.com/resource/en/datasheet/vl53l5cx.pdf}{https://www.st.com/resource/en/datasheet/vl53l5cx.pdf}} from STMicroelectronics, which is based on ToF technology.
This sensor is capable of acquiring 8$\times$\SI{8}{px} depth maps at \SI{15}{\hertz} with a \SI{313}{\milli\watt} power consumption.
In our nano-drone prototype, we employ a ready-to-use COTS board from SparkFun\footnote{\href{https://www.sparkfun.com/products/19013}{https://www.sparkfun.com/products/19013}} mounted on a \textit{Bitcraze prototyping expansion board} connected to the STM32 over the I2C bus. 
Depth maps are then forwarded over UART to the GAP8 SoC.
For distances between \SI{20}{\milli\meter} and \SI{20}{\centi\meter}, the datasheet specifies a measurement accuracy of $\pm15$\si{\milli\meter}, while from \SI{20}{\centi\meter} to \SI{4}{\meter} the error grows to $\pm11\%$.

\textbf{PULP-Frontnet:} this CNN addresses the human pose estimation aboard a nano-drone~\cite{pulp-frontnet}.
It processes monocular gray-scale 160$\times$\SI{96}{px} images and produces the relative pose of the human subject w.r.t. the drone as regression output, represented as 3D position in space $(x, y, z)$ and rotation angle w.r.t. the gravity z-axis ($\theta$).
Its training is performed in a supervised manner on real-world data collected in a mocap-equipped room.
We adopt this CNN architecture as the backbone for our models, which we extend to take advantage of depth information as described in Section~\ref{sec:models}.
In addition, we use it in our experimental evaluation and in-field tests as our SoA baseline.

\textbf{Simulator:} we employ the open-source Webots simulator to generate the training data for our CNN model.
In this fashion, we avoid error-prone and time-consuming dataset collection for the CNNs' training procedure, which would require significant effort given our double sensory sources.
We leverage the human models provided by Webots, including 3D models and walking gait animations, and the official Crazyflie 3D model and controllers provided by Bitcraze\footnote{\href{https://github.com/bitcraze/crazyflie-simulation}{https://github.com/bitcraze/crazyflie-simulation}}.
Section~\ref{sec:data-collection} describes our extensions to integrate the multi-zone ranging sensor, model its noise characteristics, and collect data while randomizing the appearance of human subjects and environments.
\section{Vision-depth Fusion}  \label{sec:system}

\begin{figure}[t]
  \includegraphics[width=\columnwidth]{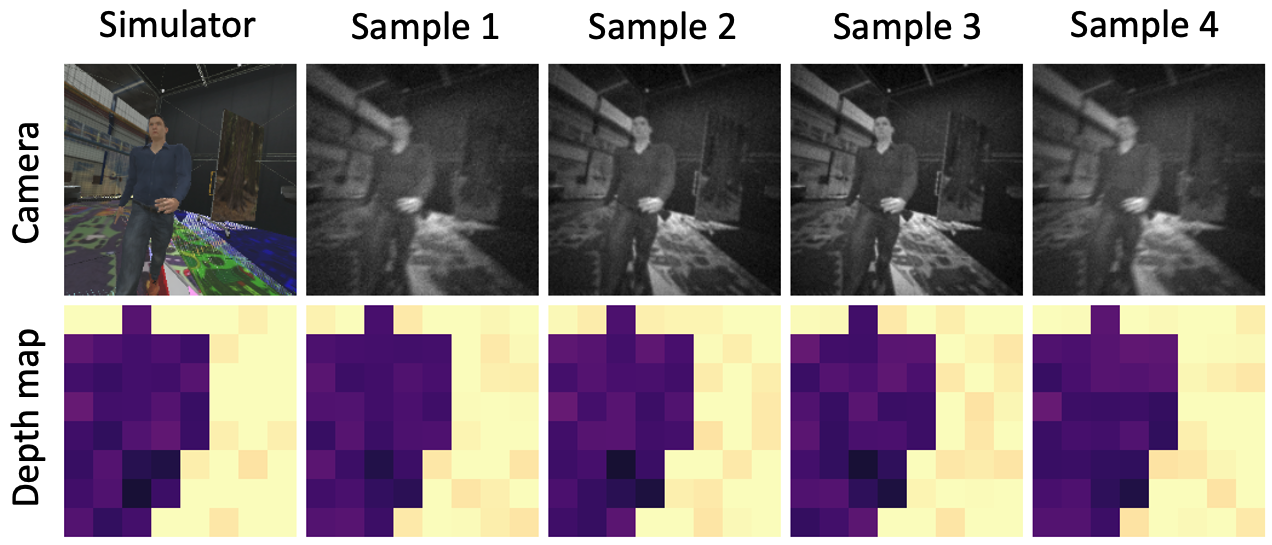}
  \caption{Four samples of image augmentations starting from the same camera image (simulator), coupled with their depth maps.}
  \label{fig:augmentations}
\end{figure}

\subsection{Neural networks fusion methods}
\label{sec:models}

The sensor fusion and pose estimation task is tackled with the CNN architecture, in Figure~\ref{fig:models}, based on a PULP-Frontnet~\cite{pulp-frontnet} backbone.
It takes two inputs: a grayscale image (160$\times$96 px) and a depth map (8$\times$8 px) produced by the multi-zone ToF depth sensor; the output consists of 4 variables: $(x, y, z)$, and the angle around the z-axis, $(\theta)$.

We compare two approaches to fuse the data from the two sensors.
The first approach includes the 8$\times$8 px depth map as one of the 64 channels in input to Block 3 of the backbone, cropped and padded to match the 10$\times$6 shape expected by the block. We name this approach \textit{mid fusion} and show it in Figure~\ref{fig:models}-B.
The second one processes the depth map with a two-layer feed-forward Multi-Layer Perceptron (MLP) network branch, in order to extract a vector of 4 features that are concatenated to the 1920 features of the vision backbone of PULP-Frontnet. We name it \textit{late fusion} MLP and report it in Figure \ref{fig:models}-C.

In addition to these two approaches, we consider two baseline methods. 
These baselines process vision and depth data with separate uni-modal sub-models, that are trained to regress our 4 output variables ($x$, $y$, $z$, and $\theta$). 
For a given test sample, we then average the outputs of the two sub-models to produce the final prediction. 
Both baselines share the PULP-Frontnet architecture represented in Figure~\ref{fig:models}-A as the vision-only sub-model, but differ in the depth-only sub-model architecture.
The \textit{Average Depth/Cam Mid} baseline, which reflects the \textit{mid fusion} option above, adopts the last part of the PULP-Frontnet network as a depth-only model: it starts from Block~3 and receives the input depth map as Figure~\ref{fig:models}-B.
On the other hand, for the \textit{Average Depth/Cam Late} baseline, the depth-only model has the architecture represented in Figure \ref{fig:models}-C.

All CNNs are trained with SGD at a learning rate of 0.001 over 100 epochs, in which each epoch consists in one pass on 150k images randomly sampled out of our 600k-image training set.
We select for testing the network that achieves the best performance on a disjoint 100k-image validation set.
The loss function used for training is the sum of the Mean Absolute Error evaluated on each predicted variable x, y, z, and $\theta$.

\subsection{Dataset collection} \label{sec:data-collection}

Our entire train and validation sets are acquired with the open-source simulator Webots, implementing a domain randomization technique~\cite{tobin2017domain}.
We iterate among 3D models of 27 persons, which vary in height, volume, and texture.
The joint poses of the person's skeleton are also randomly sampled among ten extracted from a walking gait animation.
The drone is then spawned in the environment with random position $(x, y, z)$ and orientation $(\mathit{roll}, \mathit{pitch}, \mathit{yaw})$, such that it always faces the subject.
Finally, we build the environment (background and floor) by randomly sampling among 20 textures, and we populate it with 22 random objects.

\begin{figure}[t]
  \includegraphics[width=\columnwidth]{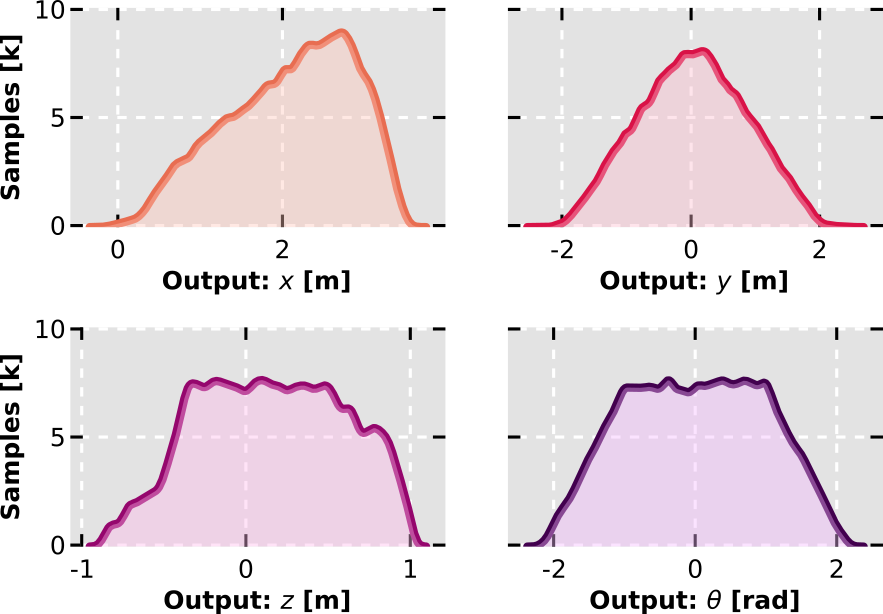}
  \caption{Sample distribution of our training dataset; each sample consists of one camera image and depth map and the corresponding ground-truth pose.}
  \label{fig:distribution}
\end{figure}

On the simulator images, we perform several photometric augmentations, such as motion blur, Gaussian blur, radial distortion, vignetting, per-pixel Gaussian noise, and brightness, to improve our networks' robustness and minimize the sim-to-real gap in the sensor data.
For each acquired image, we produce 25 augmented ones by applying all the abovementioned techniques.
While on the depth map, we apply only per-pixel Gaussian noise based on the sensor specifications (see Section~\ref{sec:background}).
In Figure~\ref{fig:augmentations}, we report a sample recorded in simulation and four different results of our augmentation pipeline.
Finally, we horizontally flip each pair of samples (camera image and depth map), with 50\% probability, and adjust the recorded relative pose accordingly.
The complete dataset recorded with the simulator consists of more than \SI{700}{\kilo\nothing} samples with output labels distributed as reported in Figure~\ref{fig:distribution}. 
Finally, we complement this training dataset with 3000 additional real-world samples (camera images, depth maps, and poses) to serve as the testing set.

\subsection{Dropout}

We train our networks with two types of dropout. 
We adopt standard dropout to drop a random subset of activations in our fully-connected layer.
In addition, we apply input dropout~\cite{input-dropout} to force the network to learn from both inputs with equal importance without overfitting on either. 
More specifically, for every training pair composed of one image and a depth map, we retain both inputs with probability $p_\text{keep}$, while with probability $p_\text{drop}$ one of the two inputs at random is masked. 
In the mid fusion network, this works by masking the corresponding feature maps in input at the 3rd block.  In the late fusion network, the features are masked at the end of Figure~\ref{fig:models}-C.

\subsection{System integration}

Since the GAP8 chip has no floating point unit and performing the computation with software-emulated floats is prohibitively expensive, we quantize weights, biases, and activations, of our CNN, to 8 bits.
The deployment phase, with the tiling for efficient memory usage, is performed by a tool that operates on an ONNX file describing the integer-quantized network and produces C code for the management of the memory used by the CNN. Since the network is particularly lightweight, the tiling and memory swapping is performed only between L1 and L2 memory, avoiding the use of the L3 memory.
The generated C code is integrated into our depth and camera acquisition pipeline onboard the AI-deck.
Since the GAP8 SoC does CNN inference on the AI-deck board, while the closed-loop control system runs onboard the Crazyflie, the GAP8 sends the estimated poses to the STM32 via UART. 
The STM32 integrates them in the closed-loop high-level controller and consequently creates a setpoint for the low-level control system.
\section{Experimental Results} \label{sec:results}
\begin{figure}[t]
  \includegraphics[width=\columnwidth]{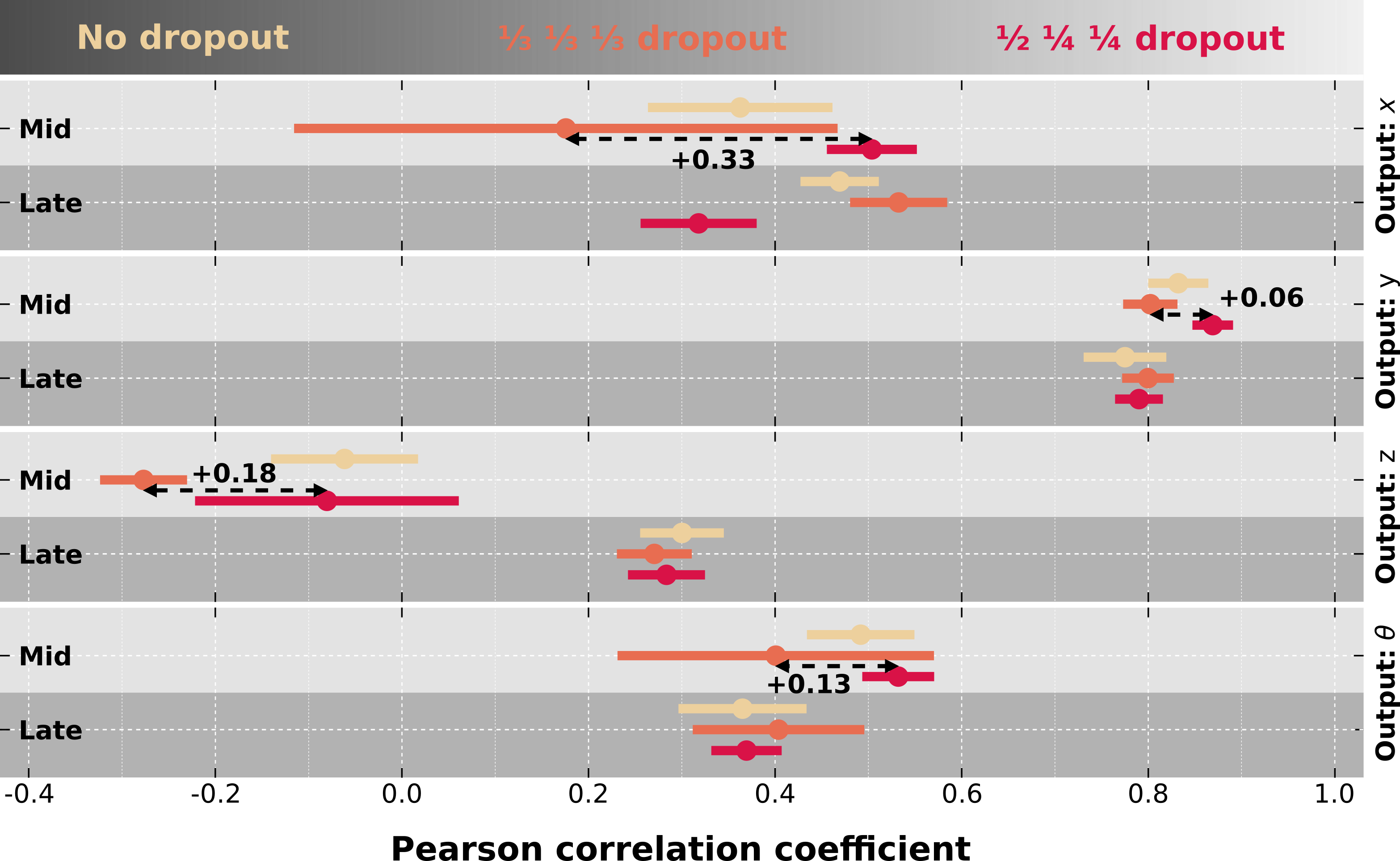}
  \caption{Average Pearson coefficient over 5 runs changing the dropout rate.}
  \label{fig:dropout}
\end{figure}

\subsection{Regression performance}

We evaluate the performance of our regression networks by reporting two metrics on the real-world test set, separately for each output variable: the Mean Absolute Error (MAE) and the Pearson correlation coefficient among predicted and ground-truth values.
The former metric measures how close the predictions are to the ground truth.  
The latter captures whether predictions are linearly correlated to the ground truth but disregards whether there are systematic biases in the predictions.
A Pearson score of 0 indicates that an increase in the ground truth value does not yield an expected increase in the predicted value, thus suggesting that the model has no predictive ability and would be useless for control.
In contrast, a Pearson score of 1 indicates a model whose outputs are perfectly linearly correlated to the ground truth, but might potentially be offset by some constant or multiplicative factor.  
When evaluating perception models for downstream control tasks, the Pearson score is indicative of the model's ability to yield stable in-field performance.

\textbf{Dropout strategy:} to select the best input dropout scheme, we test the three configurations reported in Figure~\ref{fig:dropout}: \textit{no dropout} ($p_\text{keep} = 1.0$, $p_\text{drop} = 0.0$), \textit{uniform dropout} ($p_\text{keep} = \sfrac{1}{3}$ and $p_\text{drop} = \sfrac{1}{3}$ for each input), and \textit{non-uniform dropout} ($p_\text{keep}=\sfrac{1}{2}$ and $p_\text{drop} = \sfrac{1}{4}$ for each input).
We compare these variants by considering the output variables $x$, $y$, and $\theta$, disregarding $z$ since it has limited variability in our testing data and therefore does not represent a good benchmark.
We observe that, compared to other dropout configurations, non-uniform dropout yields, on average, a lower standard deviation and higher Pearson coefficient, highlighted with the black dashed arrows in Figure~\ref{fig:dropout}.

\textbf{Fusion approach:} using non-uniform dropout, we then compare the performance of the two fusion approaches: we observe that \textit{mid-fusion} outperforms \textit{late-fusion}.
Therefore, we select the mid-fusion model with non-uniform dropout for in-field tests. 
This model yields a Pearson coefficient for the x, y, and $\theta$ variables of 0.50, 0.87, and 0.54, and a MAE of \SI{0.46}{\metre}, \SI{0.26}{\metre}, and \SI{0.30}{\radian}, respectively.

\begin{figure}[t]
  \includegraphics[width=\columnwidth]{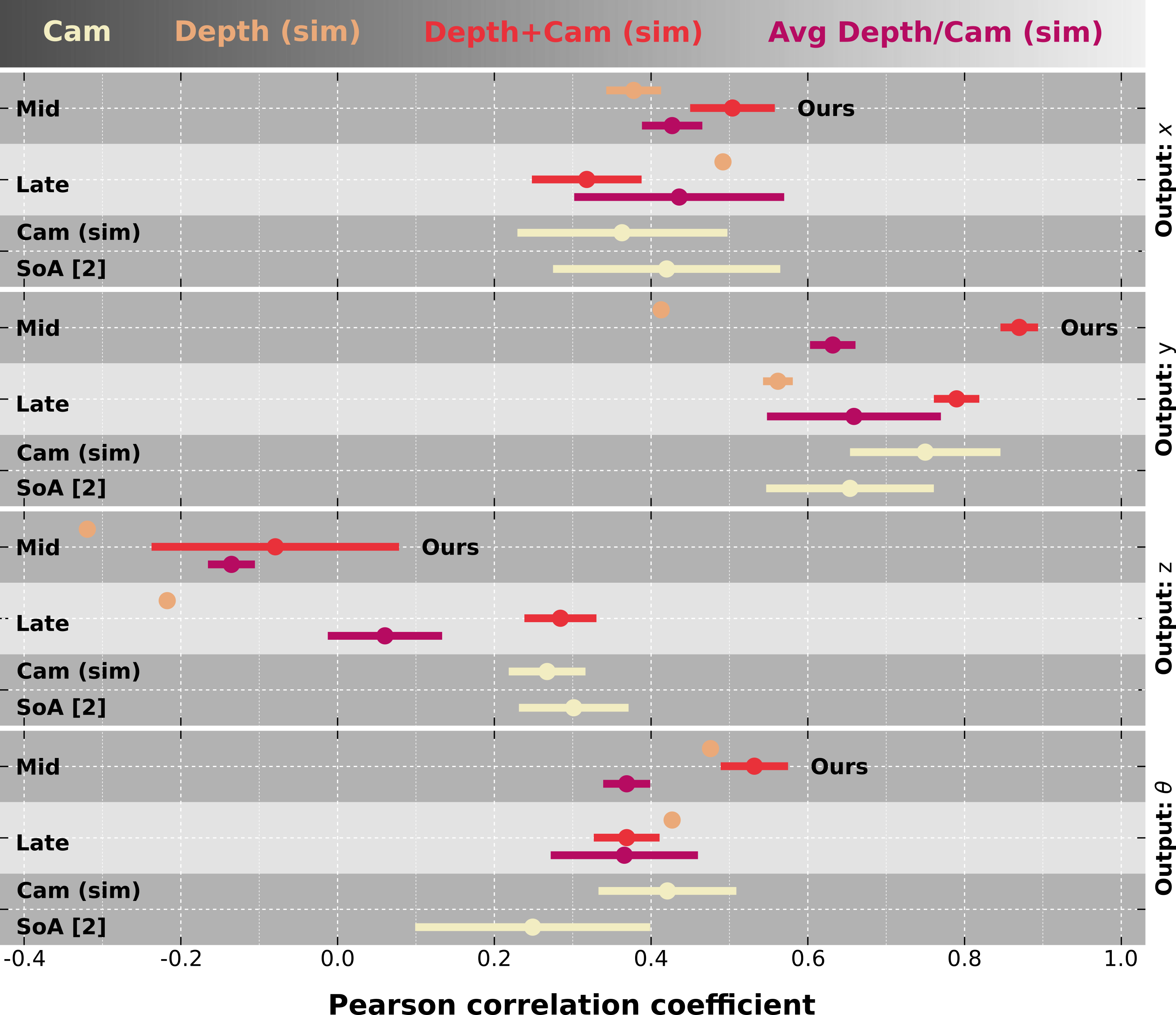}
  \caption{Means and std. deviations on 5 different training for each model.}
  \label{fig:pearson}
\end{figure}

\textbf{Comparison with baselines:} Figure~\ref{fig:pearson} reports the performance of the mid and late fusion models, named Depth+Cam (sim), against several alternatives: \textit{i}) the \textbf{SoA~\cite{pulp-frontnet}} camera-only approach, i.e., the PULP-Frontnet architecture, trained from real-world data acquired in a different lab; \textit{ii}) \textbf{Cam (sim)}, the same model trained on our simulated training set; \textit{iii}) \textbf{Depth (sim)}, a depth-only model with an architecture matching the depth branch of the mid-fusion or late-fusion models; \textit{iv}) \textbf{Avg Depth/Cam (sim)}: the average of the outputs of the two previous models.

\begin{figure}[t]
  \includegraphics[width=\columnwidth]{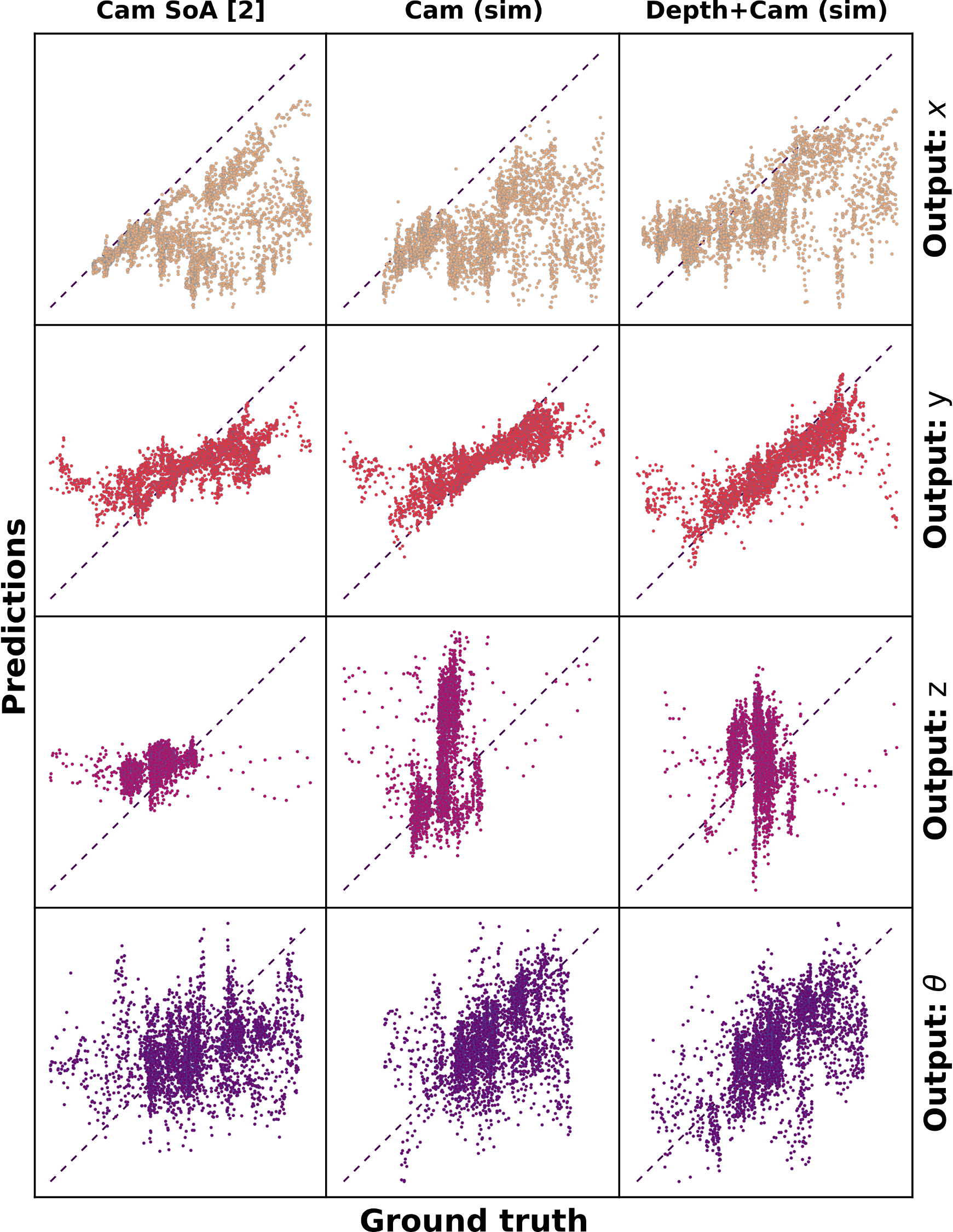}
  \caption{Predictions (y-axis) vs. ground truth (x-axis) for each system, on all outputs. Dashed diagonal lines correspond to a perfect predictor.}
  \label{fig:scatter}
\end{figure}

Our \textit{mid-fusion} approach improves the PULP-Frontnet SoA in terms of the Pearson coefficient by 30$\%$, 24\%, and 128\% on x, y, and $\theta$, respectively.
Furthermore, our approach reduces the MAE w.r.t the SoA by 41\%, 33\%, and 21\% on x, y, and $\theta$.
The \textit{Cam (sim)} network marginally outperforms the SoA on the x output variable; on the y output variable, it improves both the Pearson and the MAE metrics by 21\% and 27\%.
On the $\theta$ variable, we measure a 60\% improvement in the Pearson coefficient, while the improvement on MAE is negligible.
The baseline Avg Depth/Cam (sim) approach performs similarly to the vision-only based system trained on our simulator training set. 
Figure \ref{fig:scatter} illustrates the performance of our approach:
a substantial improvement may be observed especially on the $y$, and $\theta$ variables, and for high $x$ values, where the SoA network significantly underperforms. 

\subsection{In-field evaluation} \label{sec:in-field}

We put our system to the test in a closed-loop experiment, reproducing the test setup of our baseline~\cite{pulp-frontnet}: a human subject moves along a predefined path of increasing difficulty while the drone is autonomously controlled to stay in front of the subject, at a distance of \SI{1.5}{\meter}.
Our experiment comprises two test subjects in a motion capture-equipped environment never seen during training.
Flights are performed using three models: the camera-only SoA~\cite{pulp-frontnet}; Cam (sim); and our proposed approach Depth+Cam (sim).
We perform three flights for each combination of subject and model ($2\times3\times3=18$ flights), plus one additional flight in which the drone is controlled based on the perfect mocap position of the subject, for a total of 19 test flights. 

Table~\ref{tab:infield_results} reports the results for this experiment broken down into three groups of metrics: overall path completion, regression performance, and control performance.
We quantify path completion in terms of mean elapsed flight time and mean percentage of covered distance, terminating an experiment run as soon as the subject leaves the drone camera's field of view.
Control performance is quantified with two error metrics: the mean horizontal distance error $e_{xy}$ with respect to the desired position of the drone (i.e., \SI{1.5}{\ meter} in front of the subject); the mean absolute angular error $e_{\theta}$ of the drone orientation with respect to its desired orientation (i.e., looking towards the subject).

\begin{table}[t]
    \begin{center}
      \caption{In-field experiment results (average over 6 runs). 
      The symbol * indicates the networks trained in simulation.}
      \label{tab:infield_results}
      \resizebox{\columnwidth}{!}{
      \renewcommand{\arraystretch}{1.25}
      \begin{tabular}{lccccccc} 
        \toprule
        \multirow{2}[3]{*}{\textbf{Network}} & \multirow{2}[3]{*}{\shortstack[c]{\textbf{Flight}\\\textbf{time [s]}}} & \multirow{2}[3]{*}{\shortstack[c]{\textbf{Completed}\\ \textbf{path [\%]}}} & \multicolumn{3}{c}{\textbf{MAE}} & \multicolumn{2}{c}{\textbf{Control error}} \\
        \cmidrule(lr){4-6} \cmidrule(lr){7-8}
        & & & $x$ & $y$ & ${\theta}$  & $e_{xy}$ [m] & $e_\theta$ [rad]\\
        \midrule
        \textbf{Mocap} & 165 & 100 & 0.0 & 0.0 & 0.0 & 0.18 & 0.21\\
        \midrule
        \textbf{SoA~\cite{pulp-frontnet}} & 140 & 85 & 0.79 & 0.23 & 0.79 & 0.99 & 0.75\\
        \textbf{Cam*}  & 157 & 95 & 0.67 & 0.39 & 0.49 & 0.70 & 0.53\\
        \textbf{Depth+Cam*}  & \textbf{165} & \textbf{100} & \textbf{0.36} & \textbf{0.12} & \textbf{0.32} & \textbf{0.42} & \textbf{0.37}\\
        \bottomrule
      \end{tabular}
      }
    \end{center}
  \end{table}

We observe that the baseline PULP-Frontnet model struggles due to issues in generalization to the unseen environment and subjects, completing only 85\% of the path on average.
On the other hand, models trained in simulation exhibit higher resiliency, with the Depth+Camera model consistently completing 100\% of the path.
Regression performance, measured in terms of MAE, captures the ability of a model to estimate the subject's pose accurately.
The Depth+Camera model achieves the best absolute performance, especially on $y$ with an MAE as low as \SI{12}{\centi\meter} and less than half the MAE of the PULP-Frontnet baseline on all outputs.

These improvements directly reflect on the control performance, i.e., the accuracy of the closed-loop system in tracking the subject along the path.
Depth+Camera shows the best performance by a large margin, less than double the control error of the mocap-based flight, which represents the control error lower bound achievable with perfect sensing.
Our supplementary video shows the in-field behavior of the models, where the superior control accuracy of the Depth+Camera (sim) model is clearly noticeable.

\subsection{Discussion}
Extensive tests have been done to evaluate how the introduction of the depth sensor affects the physical behavior of the drone. 
In fact, we tested the SoA with and without the weight introduced by the sensor (not used for the pose estimation).  
On an average of six runs, the configuration with the depth sensor onboard has remarkably worse infield control performance. 
The mean position error, e$_{xy}$, is \SI{0.99}{\meter} for the configuration without the depth sensor and \SI{1.24}{\meter} for the configuration with it.
Evaluating the mean angular error, e$_{\theta}$ the performance with the depth sensor onboard deteriorates by 25\% starting from \SI{0.75}{\radian} of error of the model without the sensor onboard as reported in Table \ref{tab:infield_results}.

\begin{figure}[tb]
  \includegraphics[width=\columnwidth]{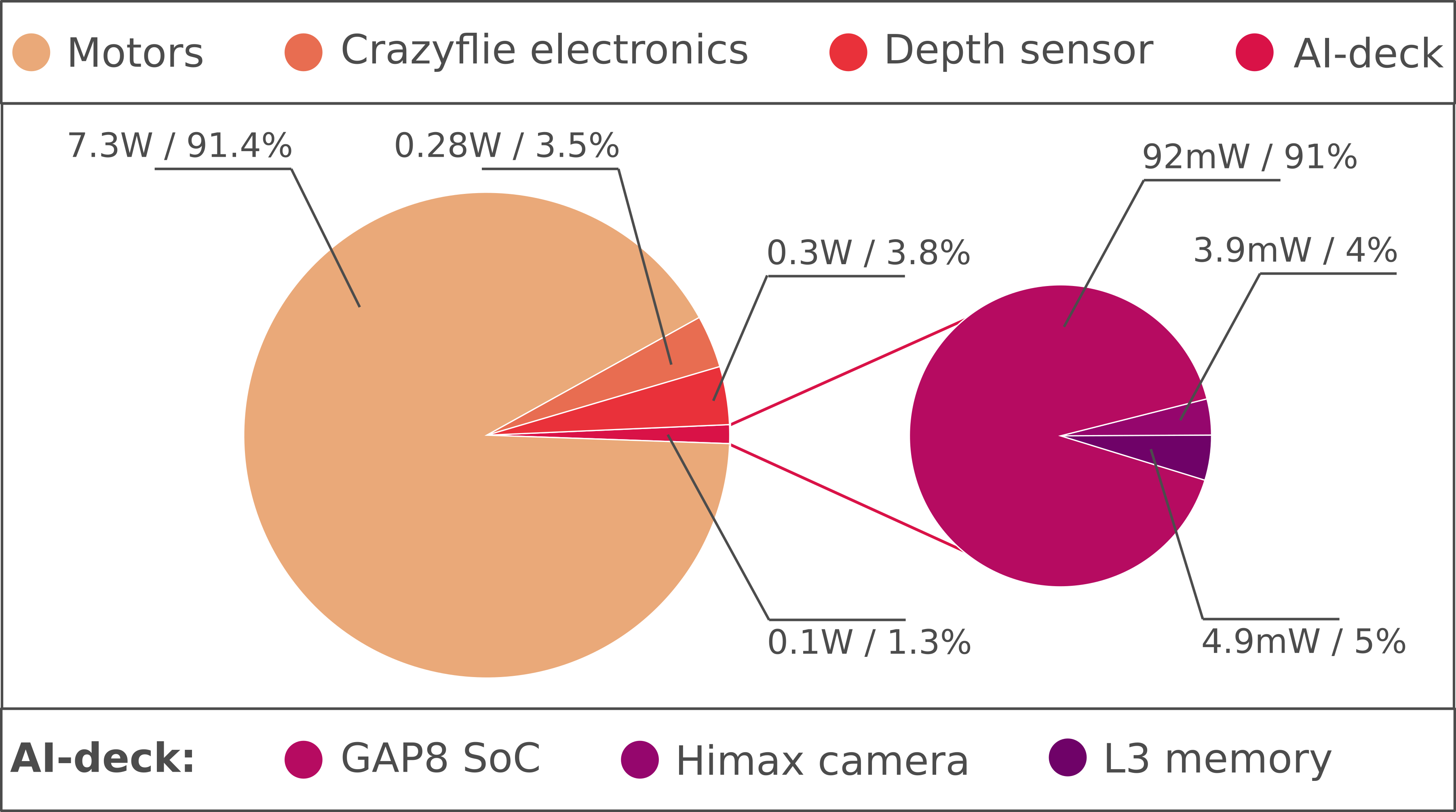}
  \caption{System power breakdown, while running the Depth+Camera model.}
  \label{fig:power_breakdown}
\end{figure}

Further, we analyze the onboard performance of our system. 
The proposed mid-fusion CNN takes advantage of the additional depth information without increasing the computational or memory requirements of the baseline CNN and thus can reach the same maximum inference throughput of 45.3fps on the GAP8 SoC.
The current depth sensor configuration, i.e., $8\times8$~px @ \SI{15}{\hertz}, results in a given depth map being fed to the CNN for roughly three consecutive inferences when running at the maximum throughput.
Future work should explore different trade-offs between depth map resolution and frame rate (up to $4\times4$~px @ \SI{60}{\hertz}) to determine how they impact our method.
Finally, we break down our system's power consumption in Figure~\ref{fig:power_breakdown}: the depth sensor accounts for an extra \SI{313}{\milli\watt} compared to the baseline, while consumption of the rest of the system remains unchanged.
Although only 3.8\% of the total power budget, the depth sensor represents 45.8\% of the power budget dedicated to sensing and computation.
Thanks to the input dropout~\cite{input-dropout} applied at training time, we expect our CNN's inference-time performance to gracefully degrade in case of missing inputs, which would enable us to selectively enable the depth sensor only part of the time to save power.
\section{Conclusion} \label{sec:conclusion}

This work presents a vertically integrated \SI{10}{\centi\meter} nano-drone system for the human pose estimation task on an ultra-constrained SoC, i.e., sub-\SI{100}{\milli\watt} compute power.
Combining a COTS 8$\times$8 multi-zone depth sensor with a low-resolution monochrome camera, we explore different CNNs to fuse these complementary inputs.
We leverage the Webots simulator to efficiently collect our multi-sensory training data.
Thanks to our dataset generation pipeline, which includes aggressive photometric augmentations, balanced label distributions, and multiple environments configurations, we deliver multiple sim-to-real models fully working in real-world testing environments.
Finally, we deploy the best models aboard our prototype nano-drone and achieve a real-time inference rate up to \SI{45}{\hertz}, within \SI{92}{\milli\watt}.
Our in-field experimental results show an improvement of 58\% and 51\% of our depth+camera system w.r.t. a camera-only SoA baseline on the horizontal and angular mean pose errors, respectively.
Finally, by employing only ready-to-use COTS components, we foster the research community to adopt this novel class of systems.

\bibliographystyle{IEEEtran}
\bibliography{IEEEabrv,biblio}

\begin{thebibliography}{10}
\providecommand{\url}[1]{#1}
\csname url@rmstyle\endcsname
\providecommand{\newblock}{\relax}
\providecommand{\bibinfo}[2]{#2}
\providecommand\BIBentrySTDinterwordspacing{\spaceskip=0pt\relax}
\providecommand\BIBentryALTinterwordstretchfactor{4}
\providecommand\BIBentryALTinterwordspacing{\spaceskip=\fontdimen2\font plus
\BIBentryALTinterwordstretchfactor\fontdimen3\font minus
  \fontdimen4\font\relax}
\providecommand\BIBforeignlanguage[2]{{%
\expandafter\ifx\csname l@#1\endcsname\relax
\typeout{** WARNING: IEEEtran.bst: No hyphenation pattern has been}%
\typeout{** loaded for the language `#1'. Using the pattern for}%
\typeout{** the default language instead.}%
\else
\language=\csname l@#1\endcsname
\fi
#2}}

\bibitem{dumbgen2022blind}
F.~D{\"u}mbgen \emph{et~al.}, ``Blind as a bat: Audible echolocation on small
  robots,'' \emph{IEEE Robotics and Automation Letters}, 2022.

\bibitem{pulp-frontnet}
D.~Palossi \emph{et~al.}, ``Fully onboard ai-powered human-drone pose
  estimation on ultralow-power autonomous flying nano-uavs,'' \emph{IEEE
  Internet of Things Journal}, vol.~9, no.~3, pp. 1913--1929, 2021.

\bibitem{DCOSS19}
------, ``An open source and open hardware deep learning-powered visual
  navigation engine for autonomous nano-uavs,'' in \emph{2019 15th
  International Conference on Distributed Computing in Sensor Systems
  (DCOSS)}.\hskip 1em plus 0.5em minus 0.4em\relax IEEE, 2019, pp. 604--611.

\bibitem{scaramuzza_alphapilot}
P.~Foehn \emph{et~al.}, ``Alphapilot: Autonomous drone racing,''
  \emph{Autonomous Robots}, vol.~46, no.~1, pp. 307--320, 2022.

\bibitem{deeppilot_2020}
L.~O. {Rojas-Perez} \emph{et~al.}, ``{{DeepPilot}}: {{A CNN}} for {{Autonomous
  Drone Racing}},'' \emph{Sensors}, vol.~20, no.~16, p. 4524, Jan. 2020.

\bibitem{uav-lidar}
Y.~Lin \emph{et~al.}, ``Mini-uav-borne lidar for fine-scale mapping,''
  \emph{IEEE Geoscience and Remote Sensing Letters}, vol.~8, no.~3, pp.
  426--430, 2011.

\bibitem{uav-rgbd}
F.~J. Perez-Grau \emph{et~al.}, ``Multi-modal mapping and localization of
  unmanned aerial robots based on ultra-wideband and rgb-d sensing,'' in
  \emph{2017 IEEE/RSJ International Conference on Intelligent Robots and
  Systems (IROS)}, 2017, pp. 3495--3502.

\bibitem{fusenet}
C.~Hazirbas \emph{et~al.}, ``Fusenet: Incorporating depth into semantic
  segmentation via fusion-based cnn architecture,'' in \emph{Computer Vision --
  ACCV 2016}, S.-H. Lai, V.~Lepetit, K.~Nishino, and Y.~Sato, Eds.\hskip 1em
  plus 0.5em minus 0.4em\relax Cham: Springer International Publishing, 2017,
  pp. 213--228.

\bibitem{10160586}
S.~Casao \emph{et~al.}, ``A framework for fast prototyping of photo-realistic
  environments with multiple pedestrians,'' in \emph{2023 IEEE International
  Conference on Robotics and Automation (ICRA)}, 2023, pp. 9083--9089.

\bibitem{jung2018perception}
S.~Jung \emph{et~al.}, ``Perception, guidance, and navigation for indoor
  autonomous drone racing using deep learning,'' \emph{IEEE Robotics and
  Automation Letters}, vol.~3, no.~3, pp. 2539--2544, 2018.

\bibitem{dronecap}
X.~Zhou \emph{et~al.}, ``Human motion capture using a drone,'' in \emph{2018
  IEEE International Conference on Robotics and Automation (ICRA)}, 2018, pp.
  2027--2033.

\bibitem{gap8}
E.~Flamand \emph{et~al.}, ``Gap-8: A risc-v soc for ai at the edge of the
  iot,'' in \emph{2018 IEEE 29th International Conference on
  Application-specific Systems, Architectures and Processors (ASAP)}.\hskip 1em
  plus 0.5em minus 0.4em\relax IEEE, 2018, pp. 1--4.

\bibitem{zhou2022swarm}
X.~Zhou \emph{et~al.}, ``Swarm of micro flying robots in the wild,''
  \emph{Science Robotics}, vol.~7, no.~66, p. eabm5954, 2022.

\bibitem{onboard-adr-survey}
L.~O. Rojas-Perez \emph{et~al.}, ``On-board processing for autonomous drone
  racing: An overview,'' \emph{Integration}, vol.~80, pp. 46--59, 2021.

\bibitem{rgbd-cnn}
M.~Schwarz \emph{et~al.}, ``Rgb-d object recognition and pose estimation based
  on pre-trained convolutional neural network features,'' in \emph{2015 IEEE
  International Conference on Robotics and Automation (ICRA)}, 2015, pp.
  1329--1335.

\bibitem{input-dropout}
S.~de~Blois \emph{et~al.}, ``Input dropout for spatially aligned modalities,''
  in \emph{2020 IEEE International Conference on Image Processing (ICIP)},
  2020, pp. 733--737.

\bibitem{nanoflownet}
R.~Bouwmeester \emph{et~al.}, ``Nanoflownet: Real-time dense optical flow on a
  nano quadcopter,'' in \emph{2023 IEEE International Conference on Robotics
  and Automation (ICRA)}, 2023, pp. 1996--2003.

\bibitem{nano-learning-vo}
S.~Chen \emph{et~al.}, ``Towards specialized hardware for learning-based visual
  odometry on the edge,'' in \emph{2022 IEEE/RSJ International Conference on
  Intelligent Robots and Systems (IROS)}, 2022, pp. 10\,603--10\,610.

\bibitem{idsia-d2d}
S.~Bonato \emph{et~al.}, ``Ultra-low power deep learning-based monocular
  relative localization onboard nano-quadrotors,'' in \emph{2023 IEEE
  International Conference on Robotics and Automation (ICRA)}, 2023, pp.
  3411--3417.

\bibitem{dewagter14-flapstereo}
C.~De~Wagter \emph{et~al.}, ``Autonomous flight of a 20-gram flapping wing mav
  with a 4-gram onboard stereo vision system,'' in \emph{2014 IEEE
  International Conference on Robotics and Automation (ICRA)}.\hskip 1em plus
  0.5em minus 0.4em\relax IEEE, 2014, pp. 4982--4987.

\bibitem{mcguire17-of}
K.~McGuire \emph{et~al.}, ``Efficient optical flow and stereo vision for
  velocity estimation and obstacle avoidance on an autonomous pocket drone,''
  \emph{IEEE Robotics and Automation Letters}, vol.~2, no.~2, pp. 1070--1076,
  2017.

\bibitem{he2015deep}
K.~He \emph{et~al.}, ``Deep residual learning for image recognition,'' in
  \emph{Proceedings of the IEEE conference on computer vision and pattern
  recognition}, 2016, pp. 770--778.

\bibitem{NIPS2017_3f5ee243}
\BIBentryALTinterwordspacing
A.~Vaswani \emph{et~al.}, ``Attention is all you need,'' in \emph{Advances in
  Neural Information Processing Systems}, I.~Guyon, U.~V. Luxburg, S.~Bengio,
  H.~Wallach, R.~Fergus, S.~Vishwanathan, and R.~Garnett, Eds., vol.~30.\hskip
  1em plus 0.5em minus 0.4em\relax Curran Associates, Inc., 2017. [Online].
  Available:
  \url{https://proceedings.neurips.cc/paper_files/paper/2017/file/3f5ee243547dee91fbd053c1c4a845aa-Paper.pdf}
\BIBentrySTDinterwordspacing

\bibitem{Liu_2022_CVPR}
C.~Liu \emph{et~al.}, ``Network amplification with efficient macs allocation,''
  in \emph{Proceedings of the IEEE/CVF Conference on Computer Vision and
  Pattern Recognition (CVPR) Workshops}, June 2022, pp. 1933--1942.

\bibitem{tobin2017domain}
J.~Tobin \emph{et~al.}, ``Domain randomization for transferring deep neural
  networks from simulation to the real world,'' in \emph{2017 IEEE/RSJ
  international conference on intelligent robots and systems (IROS)}.\hskip 1em
  plus 0.5em minus 0.4em\relax IEEE, 2017, pp. 23--30.

\end{thebibliography}

\end{document}